\documentclass[]{article}

\title{PiMPeR: Piecewise Dense 3D Reconstruction from Multi-View and Multi-Illumination Images}
\author{Reza Sabzevari$^1$, Vittori Murino$^2$, and Alessio Del Bue$^2$\\\\
	$^1$ Robotics and Perception Group, University of Zurich, Switzerland\vspace{.2cm}\\
	$^2$ Pattern Analysis and Computer Vision (PAVIS),\\ Italian Institute of Technology, Genova, Italy}

\usepackage{graphicx}
\usepackage{amsmath}
\usepackage{amssymb}
\usepackage{hyperref}
\usepackage{array}

\def\mat#1{\mathchoice{\mbox{\boldmath$\displaystyle\tt#1$}}
{\mbox{\boldmath$\textstyle\tt#1$}}
{\mbox{\boldmath$\scriptstyle\tt#1$}}
{\mbox{\boldmath$\scriptscriptstyle\tt#1$}}}

\def\vec#1{\mathchoice{\mbox{\boldmath  $\displaystyle\bf#1$}}
{\mbox{\boldmath  $\textstyle\bf#1$}}
{\mbox{\boldmath  $\scriptstyle\bf#1$}}
{\mbox{\boldmath  $\scriptscriptstyle\bf#1$}}}

\newlength{\colwidth}
\begin{document}

\maketitle

\begin{abstract}
In this paper, we address the problem of dense 3D reconstruction from multiple view images subject to strong lighting variations.
In this regard, a new piecewise framework is proposed to explicitly take into account the change of illumination across several wide-baseline images.
Unlike multi-view stereo and multi-view photometric stereo methods, this pipeline deals with wide-baseline images that are uncalibrated, in terms of both camera parameters and lighting conditions.
Such a scenario is meant to avoid use of any specific imaging setup and provide a tool for normal users without any expertise.
To the best of our knowledge, this paper presents the first work that deals with such unconstrained setting.
We propose a coarse-to-fine approach, in which a coarse mesh is first created using a set of geometric constraints and, then, fine details are recovered by exploiting photometric properties of the scene.
Augmenting the fine details on the coarse mesh is done via a final optimization step.
Note that the method does not provide a generic solution for multi-view photometric stereo problem but it relaxes several common assumptions of this problem.
The approach scales very well in size given its piecewise nature, dealing with large scale optimization and with severe missing data. 
Experiments on a benchmark dataset \textit{Robot data-set} show the method performance against 3D ground truth. 

\end{abstract}

\section{Introduction}
\label{sec:intro}

This paper addresses the problem of dense reconstruction from several uncalibrated multi-view images that are subject to lighting variation.
Such a dense reconstruction pipeline is meant to deal with a off-the-shelf hardware setup---both in terms of image acquisition and processing---in a reasonable amount of time.
The proposed method is based on two fundamental reconstruction methods and accordingly consists of two main stages: first, using \textit{Structure from Motion} (SfM) to recover the coarse geometry of the scene/shape, and second using \textit{Photometric Stereo} (PS) to recover fine surface details exploiting photometric properties of the surface.
In this way, the method can take maximum advantage of geometric and photometric scene properties captured by multi-view and multi-illumination images.
For this reason, we call our proposed pipeline as PiMPeR stands for \textit{Piecewise Multi-view Photo-geometric dense Reconstruction}.
A sparse 3D point cloud is obtained using SfM by tracking and matching a set of sparse feature points on the image sequence.
This step also provides the camera positions, which will be used in the further steps of the pipeline.
Such a 3D point cloud will be too sparse, because very few feature tracks will survive among the images under drastic variation of lighting conditions.
The coarse 3D mesh is defined by first projecting the point cloud on one of the images and performing 2D Delaunay triangulation, and then applying the same triangulation on the point cloud.
Such a 3D mesh allows us to decompose the images into a set of image patches to simplify the establishment of multi-view correspondences and it  further helps to assemble the fine-detailed surface patches to form a dense surface. 
Once the multi-view images are decomposed, the corresponding image patches are registered on a template patch and the PS is used to recover dense surfaces for all the patches.
This requires adoption of classic PS methods and their reformulation in a piecewise framework, which consequently requires an integration step to collect all surface patches to a uniform one.

\subsection{Comparisons with related work}
\label{sec:related_work}

Recent advancements in rigid SfM have pushed the boundaries in different aspects, such as large scale 3D modelling in urban areas~\cite{Agarwal:etal:2010}, real-time dense 3D reconstruction \cite{Newcombe:Davison:2010} and, most recently, chronological scene reconstruction \cite{matzen2014scene}. 
These methods have either assumptions on fixed lighting conditions in real-time scenarios or use a tremendous amount of images to find a much smaller number of points that can be tracked across the images.
However, in realistic imaging conditions, the scene illumination is usually subject to change.
Moreover, the variation of the lighting conditions is fundamental to accurately reconstruct parts of the scene with homogeneous texture.
Recently, Pizzoli et al. \cite{pizzoli2014remode} and Engel et al. \cite{engel2014lsd} showed that only high gradient textures could be used as reliable observations of the scene for 3D reconstruction, and that, to recover 3D position of the scene parts with homogeneous textures, strong prior information such as surface smoothness is essential.

Differently, we present a novel method that explicitly takes into account changing illumination and uses them to obtain dense 3D reconstructions in an uncalibrated and multi-view setup. 
In order to tackle this problem, the classic formulation of PS is adapted in a piecewise framework on multi-view image patches registered on a template patch with a considerable amount of missing data.
The dense matching of multi-view image patches is possible by multi-view constraints imposed by a set of sparse image points that could be reliably tracked over the images.

To this end, our approach proposes a multi-view piecewise decomposition of the images that permits us to solve locally both the image pixel correspondence and the photometric stereo problems. 
In particular, we are able to deal with the most difficult case of wide-baseline images, where given both lighting and geometric distortions, we have image patches with strong scale changes, self-occlusion, cast-shadows, and strong specularities. 
Previous approaches, instead strongly rely on small baselines or custom setups in order to establish reliable correspondences between pixels in different frames \cite{zhang2003shape, lim2005passive, joshi2007shape, hernandez2008multiview}. 
Other geometric methods that do not exploit illumination variation rely on large datasets of images and require powerful processing resources \cite{agarwal2011building}.
In such a scenario, our proposed framework is a step forward towards dense reconstruction from a few uncalibrated images with arbitrary camera motion and lighting conditions. 

Factorization-based methods for PS have certainly provided efficient closed-form solutions without using specific equipment such as structured light. 
It is based on the fact that a set of images taken from a static view and subject to varying lights lies in a certain subspace. 
Hayakawa \cite{Hayakawa:1994} first made evident such constraints assuming a Lambertian surface and a single light source. 
Basri et al. \cite{basri2007photometric} used a more descriptive photometric model based on a spherical harmonic representation of lighting variations. 
These classic methods in PS, which do not have any depth assumption, always tie with the bas-relief ambiguity~\cite{belhumeur1999bas}. 
On the contrary, Shi et al. \cite{shi:etal:2010} perform an automatic radiometric calibration by identifying a new set of constraints that can solve for the Generalised Bas-Relief (GBR) transformation. 
Finally, Papadhimitri and Favaro \cite{Papadhimitri:Favaro:2013} proposed a closed-form solution based on the identification of a set of local diffuse reflectance maxima which provides enough constraints to solve for the GBR ambiguity. 

Another viable option to obtain reliable and detailed 3D surfaces is to use a structured pattern for illuminating the object in a controlled environment \cite{scharstein2003high}. 
These active systems have led to a high number of custom solutions \cite{Salvi:etal:2004} that require a laboratory setup and accurate calibration of the devices. 
Hernandez et al. \cite{hernandez:etal:2007}, instead, use a less restrictive calibrated setup with three not collinear colored lights in a dark room with surfaces that also need to be photometrically calibrated. 
Recently, Anderson et al. have extended this approach to arbitrary colored surfaces \cite{Anderson:etal:2011}. 
Finally, with a less constrained setup, the method of Park et al. \cite{Park:etal:2013} considers a coarse-to-fine approach that requires only a stereo pair.

More related to our approach, Multi-View Photometric Stereo (MVPS) use both geometric and photometric constraints to solve the problem in multi-view setups but using short baselines \cite{zhang2003shape}. 
Similarly, Lim et al. \cite{lim2005passive} use multiple views to generate a coarse planar surface of the object as a 3D triangulated mesh based on recovered sparse 3D points. 
Then, they perform PS in an iterative way for each triangle to recover the dense surface and to align it with the recovered 3D points. 
They also assume the lighting conditions to be fixed and that they can reconstruct the objects having small baselines only.
Joshi and Kriegman \cite{joshi2007shape} define the multi-view photometric algorithm as the optimization of multi-view and photometric matching costs using a graph-cut based approach. 
The main drawback of the method is an inherent fronto-parallel assumption that leads to local rank constraints in the images.

In general, a piecewise formulation helps to deal with changes of both geometrical and photometric components over time by simplifying the multi-view correspondence problem. 
Piecewise methods have recently demonstrated their strength for solving SfM problems of increasing complexity (see \cite{Varol:ICCV2009},  \cite{Taylor:etal:2010} and \cite{Fayad:etal:2010}).

\subsection{Contribution}
Considering the problem that is addressed in this paper, one of the most challenging scenarios for dense 3D reconstruction has been targeted.
We do not use a specific imaging setup and we consider images from uncalibrated cameras with arbitrary lighting conditions and view-points. 
Unlike other methods (i.e. \cite{engel2014lsd, agarwal2011building}), that require many images and huge amounts of processing resources to create fairly dense point clouds, we use a few images and the processing capacities available on normal PCs to obtain dense surfaces.

Differently from \cite{joshi2007shape,lim2005passive,zhang2003shape}, we deal with a wide-baseline scenario where the images can be taken after consistent camera motion, which is the current weakness of MVPS algorithms. 
Using a piecewise formulation allows us to solve for the MVPS problem using local estimates of the photometric properties of the shape, which results in efficiently by solving several small optimization problems rather than a global one. 
Using piecewise photometric stereo (PPS) we can deal with local shape ambiguities using our local-to-global surface patches alignment.
 
\subsection{Structure of the Paper}

The rest of the paper is structured as follows. 
Section~\ref{sec:PiMPS} describes PiMPeR,our pipeline for photo-geometric dense reconstruction. 
It details all the steps of our pipeline: the generation of a coarse mesh, the recovery of detailed surface patches and the specification of the optimization problem that is used to assemble the detailed surface patches on the coarse mesh.
Experiments in Section~\ref{sec:experiments} present the 3D reconstruction obtained by our approach and, finally, Section~\ref{sec:conclusions} draws the conclusions and the path for future research potentials.

\begin{figure*}[tb]
%\captionsetup{aboveskip=5pt}
%\captionsetup{belowskip=-10pt}
\begin{center}
\includegraphics[trim = 4mm 45mm 2mm 45mm, clip=true, width=\linewidth]{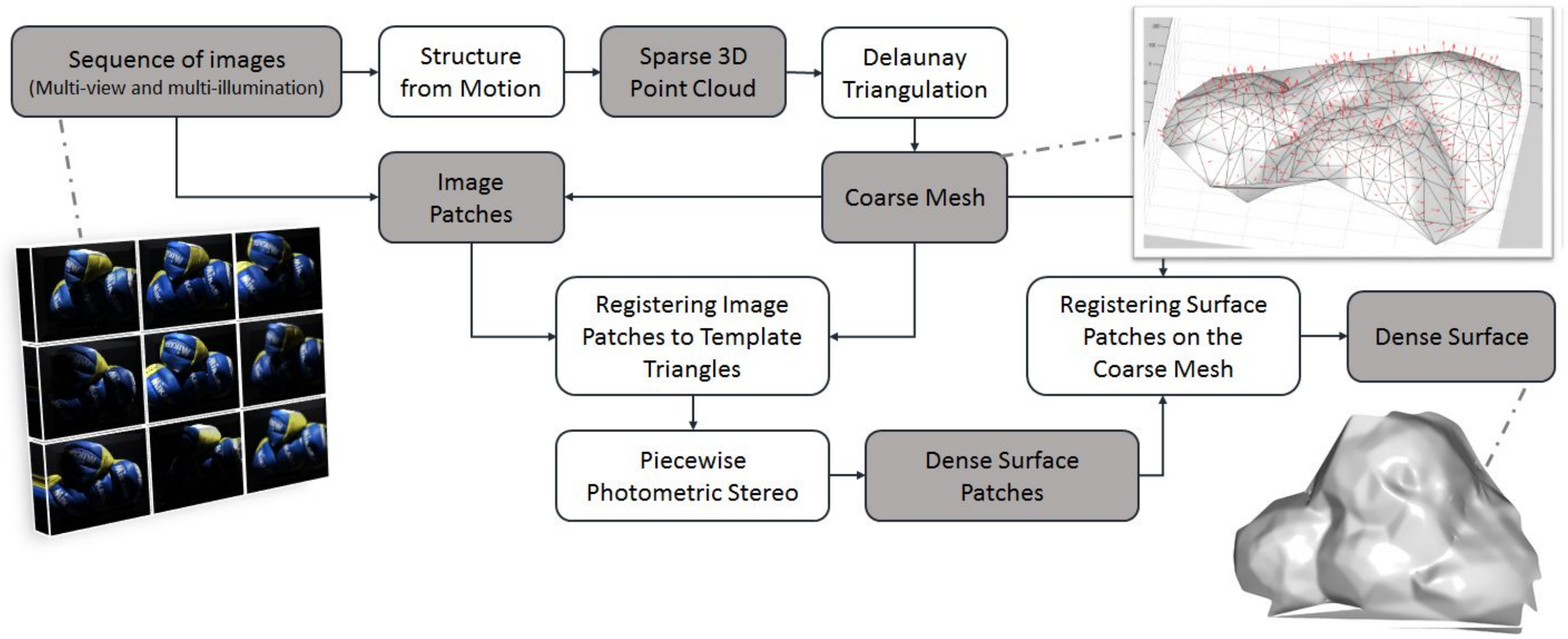}
\end{center}
\caption{The PiMPeR pipeline: input sequence contains images with different viewpoints and arbitrary lighting; the coarse 3D mesh generated by SfM and then projected onto the images; the input sequence is decomposed to triangular image patches; triangular image patches are registered on a template triangle, which is the corresponding facet on the coarse mesh; detailed surface patches are recovered using piecewise photometric stereo; the complete surface is obtained by globally registering all the surface patches on the coarse mesh.}
\label{fig:PiMPS_pipeline}
\end{figure*}

\section{The PiMPeR Pipeline}
\label{sec:PiMPS}

Our approach combines both multi-view and photometric constraints to obtain a complete 3D reconstruction of the object together with its photometric properties. 
The complete pipeline is sketched in Figure \ref{fig:PiMPS_pipeline} and we briefly resume each stage before describing them in more detail in the next sections. 
Note that the input to our algorithm is only a set of uncalibrated images, object albedo and lighting directions are not given as well. 
First we create a coarse 3D mesh using SfM by tracking/matching a set of feature points in the image sequence. 
Given the adverse lighting conditions, we account for large percentages of missing data in the tracked 2D features by adopting robust SfM methods.  next stages. 
Each triangle of the mesh corresponds to a set of image patches.
These image patches are first registered to a common image template and then the corresponding surface patches are reconstructed using a local photometric stereo method. 
At the final stage, the geometric constraints on the surface---defined by the 3D mesh---are used to assemble the surface patches, taking advantage of the proposed photo-geometric alignment.

\subsection{Mesh Generation}
\label{ssec:pimps:mesh generation}

The PPS reconstruction requires an initial partition of the images into a set of patches.
For this reason, a 3D mesh is computed by performing multi-view 3D reconstruction from a set of 2D coordinates tracked/matched in the image sequence. This task can be efficiently done using a robust method for SfM that can cope with a large number of missing data in the measurements followed by a 3D mesh generation stage (see \cite{sabzevari2012piecewise}). 
In more detail, after recovering the 3D positions belonging to $p$ image correspondences via a rigid SfM and storing them in matrix $\mat S_{sfm}$, a Delaunay triangulation is computed to generate a coarse 3D mesh having triangular image patches.
Such a set $\mathcal T$ of $m$ triangles is given by:
\begin{equation}
\label{eqn:set of 2D triangles}
	\mathcal T = 
	\left\{\vec	t_1 \; \ldots \; \vec t_m	\right\}
	\; , \hspace{.4cm}
	\vec t_{\mu} \in \left\{	1 \; \ldots \; p	\right\}^3
	\;\; \vert \; 
	\mu =	1 \; \ldots \; m	
\end{equation}
where $\vec t_{\mu}$ contains the $3$ indices identifying the $3$ vertices of the triangle $\mu$. 
Now, the vertices of triangle $\mu$ for each image frame $g$, $(g = 1 \dots f)$, form a $2 \times 3$ matrix $\mat W_{g\mu}$. 
Likewise, the 3D triangle vertices are stored in a $3 \times 3$ matrix $\mat Z_{\mu}$. 
This defines two sets of matrices that form the 3D mesh as $\mathcal S = \left\{\mat Z_{\mu} \right\}$  and the 2D mesh at each frame $g$ as $\mathcal W_g = \left\{	\mat W_{g\mu} \right\}$.

In other words, the triangulation is done on 2D image points of an arbitrary view (i.e., the frontal view) and then the resulted point indices that define the 2D triangulated mesh are used to triangulate the 3D points.

This mesh is then used for three purposes: decomposing the multi-view image frames into triangular image patches, assembling the recovered triangular surface patches in global 3D shape, and correcting the local depth ambiguities.

\subsection{Multi-View Image Patches Decomposition}
\label{ssec:pimps:image decomposition}

Once the mesh is generated, we proceed on decomposition by partitioning the image sequences into pieces that will be reconstructed using a local photometric stereo algorithm.

Consider an image as a matrix $\mat I_g$ of size $h\times w$. 
To perform the decomposition, the 3D mesh $\mathcal S$ is projected onto the image frames $\{\mat{I}_g\}$, $g \in [1 \dots f ]$, thus, providing $m \times f$ triangles. 
A triangular patch is represented as a set of $h \times w$ binary matrices $\mat{B}_{g\mu}$. 
Using such binary masks, pixels included in the triangle $\mu$ can be extracted from each image frame.
In order to ease the following steps of 3D patch alignment (Sec. \ref{ssec:pimps:patch alignment}), we enlarge the binary masks having a set of overlapping pixels with their immediate neighbouring patches. 

Let $\mat W_{g\mu}$ be the projection of triangle $t_\mu$ on image frame $g$ and
consider the enlargement of triangle $t_{\mu}$ such that the new 2D vertices on the image plane define a bigger triangle in which the new edges are parallel to the original ones.
This enlarged triangle is defined as $\tilde t_{\mu}$ with 2D vertices denoted by $\tilde{\mat W}_{g\mu}$.
The conversion from 2D vertices $\tilde{\mat W}_{g\mu}$ to an image binary mask is made using the three barycentric coordinates $\alpha_{gk}^{(\mu)}$, $\beta_{gk}^{(\mu)}$ and  $\gamma_{gk}^{(\mu)}$. 
Under this coordinate system, a pixel $k$ that belongs to triangle $\tilde{t}_{\mu}$ projected on frame $\mat I_g$ must satisfy the following condition:

\begin{equation}
\label{eqn:pixels in triangle}
\alpha_{gk}^{(\mu)} + \beta_{gk}^{(\mu)} + \gamma_{gk}^{(\mu)} = 1,
\end{equation}
where
\begin{equation}
\label{eqn:barycentric indices}
k \in \left \{ 1, \ldots, b \right \} \; , 
\hspace{.3cm}
\mathcal K_{g\mu} = \left \{ k \; : \; \alpha_{gk}^{(\mu)} + \beta_{gk}^{(\mu)} + \gamma_{gk}^{(\mu)} = 1 \right \}
\end{equation}
with $\mathcal K_{g\mu}$ being the set of indices for pixels belonging to triangle $\tilde{t}_{\mu}$ on frame $g$.
Using this formalization, the binary mask for the projection of triangle $\tilde{t}_{\mu}$ on frame $g$ is defined as a $h \times w$ matrix:

\begin{equation}
\label{eqn:binary mask}
\mat B_{g\mu} = 
\begin{bmatrix}
	B_{g1}^{(\mu)} \; \ldots \; B_{g(b-h)}^{(\mu)} \\
	\ddots \\
	B_{gh}^{(\mu)} \; \ldots \; B_{gb}^{(\mu)} 
\end{bmatrix}_{h \times w} 
\; , \hspace{.2cm}
B_{gi}^{(\mu)} = \begin{cases}
0 & i \notin \mathcal K_{g\mu}\\
1 & i \in \mathcal K_{g\mu}
\end{cases}.
\end{equation}
So, every image patch is given as:
\begin{equation}
\label{eqn:patches}
	\mat{J}_{g\mu}= \mat B_{g\mu} \odot \mat I_g,
\end{equation}
where operator $\odot$ stands for the element-wise multiplication.

\subsection{Multi-View Image Pixel Registration}
\label{ssec:pimps:multiview pixel registration}

Each triangle on the coarse 3D mesh corresponds to a set of image patches that are obtained by projecting that particular triangle on the images.
Given the camera (or object) motion, a set of image patches with different shapes and sizes will be obtained for every triangular mesh facet. 
This fact introduces a challenge, because pixel-to-pixel image patch correspondence has to be solved.
Moreover, since the patches may consistently vary their sizes, some patches might have missing pixels with respect to their corresponding reference patch. 

For this reason we need a registration process that aligns all the corresponding image patches in different views.
To that end, for each triangle $\tilde{t}_{\mu}$ we define a template and register all image patches to their corresponding templates.
Consider $\tilde{\mat Z}_{\mu}$ as a $3\times 3$ matrix including the 3D coordinates of enlarged triangle $\tilde{t}_{\mu}$ from the coarse 3D mesh $\mathcal S$. 
Such 3D triangles have a normal $\vec N_f^{\mu}$ aligned to be fronto-parallel to the image plane \cite{Horn:1987} (i.e., aligned to $ \vec N_I = [0 \;\; 0 \;\; 1]^\top$). 
The template triangle $\mat{\bar{Q}}_{\mu}$ is obtained by estimating a rotation matrix $\mat R_{\mu}$ that transforms the corresponding mesh facet and then projects the rotated triangle on an arbitrary image plane, such that:
\begin{equation}
\label{eqn:flat individually}
\mat{\bar{Q}}_{\mu} = 
\begin{bmatrix}
	1 \:\: 0 \;\; 0 \\ 0 \;\; 1 \;\; 0
\end{bmatrix}
\cdot \mat{R}_{\mu} \cdot \tilde{\mat Z}_{\mu}.
\end{equation}
Hence, $\mat{\bar{Q}}_{\mu}$ is a $2 \times 3$ matrix containing the $3$ coordinates of the template triangle as:

\begin{equation}
\label{eqn:flattened triangle}
	\mat{\bar{Q}}_{\mu} = \left[ \vec{\bar{q}}_{\mu 1} \;\; \vec{\bar{q}}_{\mu 2}\;\; \vec{\bar{q}}_{\mu 3} \right]. 
\end{equation}

A practical solution for registering multi-view triangular image patches to the corresponding templates is given by the barycentric coordinate system. 
Consider pixel $k$ from image frame $\mat I_g$ belonging to triangle $\tilde{t}_{\mu}$. 
This pixel coordinate can be determined as a 3-element barycentric coordinate:

\begin{equation}
\label{eqn:barycentric cord vector}
\vec \lambda_{gk}^{(\mu)} = 
\begin{bmatrix}
	\alpha_{gk}^{(\mu)}  &
	\beta_{gk}^{(\mu)}  &
	\gamma_{gk}^{(\mu)}
\end{bmatrix}^\top.	
\end{equation}

\begin{figure}[t]
\begin{center}
  \includegraphics[trim = 1mm 8mm 1mm 8mm, clip=true, width=0.4\linewidth]{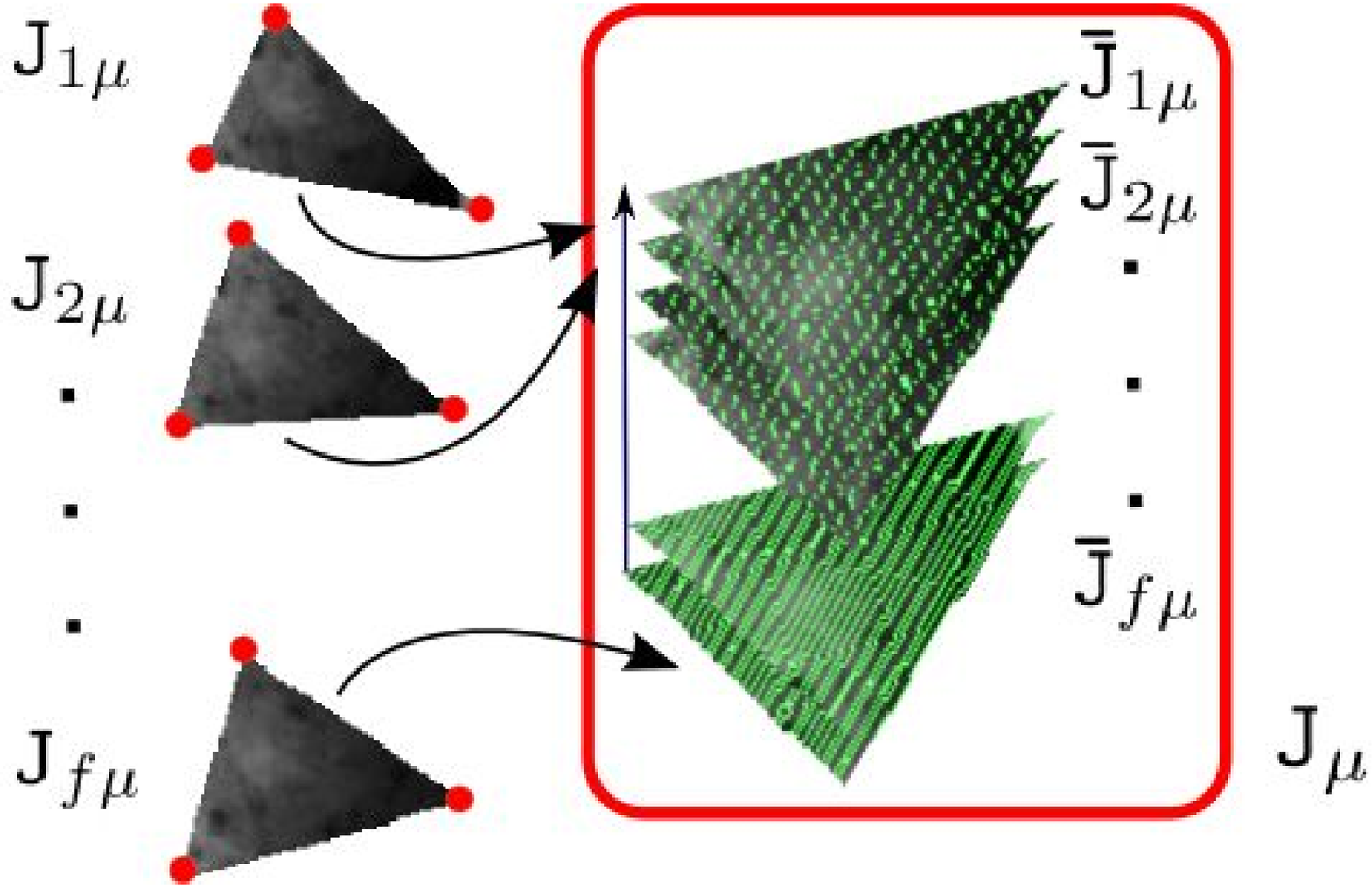}
\end{center}
\caption{Multi-view patches in different frames are projected to the corresponding template creating a stack of images where pixel correspondences are given. Green pixels show the missing values after registering multi-view image patches to the template triangle.}
\label{fig:patch_registration}
\end{figure}

In this way, every pixel $k$ belonging to triangle $\tilde{t}_{\mu}$ in all image frames can be mapped to a single position (with respect to barycentric coordinates) in the template triangle given by $\mat{\bar{Q}}_{\mu}$.
This piecewise planar mapping gives the pixel-to-pixel correspondences across all the views.
Notice that, depending on the view, the number of pixels mapped into the template may be smaller than the overall size of the template triangle. 
This might introduce a consistent number of missing pixels to the registered triangles. 
Fig.~\ref{fig:patch_registration} shows the missing pixel values for different registered image patches as green points. 
After registering all the triangles $\tilde{t}_{\mu}$ from image frames $1 \ldots f$ to the corresponding template, we can form the multi-view intensity matrix such as: 

\begin{equation}
\label{eqn:registered patches}
\mat{J}_{\mu}=
\begin{bmatrix}
vec(\bar{\mat J}_{1\mu})^\top \\ \vdots\\ vec(\bar{\mat J}_{f\mu})^\top
\end{bmatrix}_{f\times b},
\end{equation}
where vector $vec(\bar{\mat J}_{g\mu})$ ($g = 1 \dots f$) is the column-wise vectorization of registered image patch $\bar{\mat J}_{g\mu}$.
Note that, registered image patches, i.e., $\bar{\mat J}_{g\mu}$, are $h \times w$ matrices with $\left|{\mathcal K_{\mu}}\right|$ non-zero values, where $\mathcal K_{\mu}$ is the set of pixel indices in $\bar{\mat J}_{g\mu}$ that have non-zero values and represent the registered triangle $\tilde t_{\mu}$ (similar to Eq. (\ref{eqn:barycentric indices})).

\subsection{Surface Patch Reconstruction}
\label{ssec:pimps:pps}

After registering all the corresponding multi-view image patches, it is possible to compute the photometric parameters (albedo, surface normals, and lighting directions) from the matrix $\mat{J}_{\mu}$ for each multi-view image patch. 
A grayscale pixel $x_{g i}^{(\mu)} \in \mat{J}_{\mu}$ (pixel $i$ from frame $g$ belonging to triangular patch $\mu$) can be defined in terms of a normal to the surface $\vec{n}_{\mu i}$, the albedo $\rho_{\mu i}$ at that pixel position and the lighting direction $\vec{l}_{\mu g}$ such that:

\begin{equation}
\label{eqn:PPS single pixel}
\begin{array}{c}
	x_{g i}^{(\mu)} = \vec{l}_{\mu g}^\top \: \: \rho_{\mu i}  \begin{bmatrix} 1 \\ \vec{n}_{\mu i} \end{bmatrix} 
\end{array}
\end{equation}
where $\vec{l}_{\mu g} \in {\mathbb R}^{4}$, $\rho_{\mu i} \in {\mathbb R}$, $\vec{n}_{\mu i} \in {\mathbb R}^{3}$ with the non-linear constraint $\vec{n}_{\mu i}^\top \; \vec{n}_{\mu i} = 1$. 
Rewriting Eq. (\ref{eqn:PPS single pixel}) for all the pixels in patch $\mu$ and all the frames, according to \cite{basri2007photometric} yields:

\begin{equation}
\label{eqn:piecewise_PS}
\mat J_{\mu}=
\begin{bmatrix}
	\vec{l}_{\mu 1}^\top\\ \vdots \\ \vec{l}_{\mu f}^\top
  \end{bmatrix}
  \cdot
  \begin{bmatrix}
		\begin{bmatrix}
			\rho_{\mu 1} \\ \vec{n}_{\mu 1} 
		\end{bmatrix}
		\;\; \ldots \;\; 
		\begin{bmatrix}
			\rho_{\mu b} \\ \vec{n}_{\mu b} 
		\end{bmatrix}
  \end{bmatrix}
  =\mat L_{\mu} \cdot \mat N_{\mu},
\end{equation}
where the $f \times 4$ matrix $\mat L_{\mu}$ contains the collection of lighting directions for $f$ frames, while the $4 \times b$ matrix $\mat N_{\mu}$ contains the surface normal vectors and albedo values.

However, the previous multi-view registration step has produced a matrix $\mat J_{\mu}$ with missing values and thus any closed-form solutions such as \cite{basri2007photometric} cannot be applied to reconstructing the surface patch. % is not a viable options if the pixel values in the matrix $\mat J_{\mu}$ are 
To deal with this kind of missing data, we first define a binary mask for each patch $\mu$ as a matrix $\mat D_{\mu}$, containing zeros for the elements corresponding to missing values. 
As an additional source of missing data, we also consider extremely dark and saturated pixels, since they do not provide relevant information about the photometric properties of the shape. 
Thus, the estimation problem can be formalized as an optimization that solves for photometric properties of a single patch $\mu$ such that: 

\begin{equation}
\label{eqn:PS missing values}
\begin{array}[t]{ll} 
\min_{\mat L_{\mu}, \mat N_{\mu}} & \left\| \mat D_{\mu} \odot ( \mat J_{\mu} - \mat L_{\mu} \: \mat N_{\mu})  \right\|^2 
\\ \mbox{subject to} & \mat N_{\mu} \in {\mathcal N} 
\end{array},
\end{equation}
where 
$$ {\mathcal N} = \left\{ \rho \: [ 1 \:\: \vec n^\top ]^\top \:\: : \:\: \rho \in {\mathbb R}, \:\: \vec n \in {\mathbb R}^3, \: \vec n^\top \vec n =1  \right\} $$
is the \emph{photometric manifold} representing the non-linear constraint given by the $1^{st}$ order spherical harmonic approximation, as in Eq. (\ref{eqn:PPS single pixel}).
This bilinear optimization problem with manifold constraints can be solved using a general purpose solver, such as BALM \cite{del2012bilinear}.
Notice that we solve it for the surface normals associated to each pixel but not for the overall 3D surface.
Therefore, a further integration step is required to recover the final 3D surface from the surface normals \cite{frankot1988method}. 
This step provides the 3-vector $\vec{\tilde{s}}_{\mu i}$ related to pixel $x_{i}^{(\mu)}$ that represents the 3D surface point $i$ belonging to triangle $\mu$. 

Considering all triangular patches, $m$ small PS problems have to be solved. 
This piecewise formulation for photometric stereo entails several advantages with respect to previous global approaches. 
Each patch has been associated with an individual photometric model rather than a single global model, as used in classical approaches. 
In this way, the piecewise formulation may grasp more complex lighting effects.
Moreover, it is intrinsically more efficient and highly parallelizable, since reconstructing the surface in patches is computationally faster than reconstructing the global surface.  

The next Section presents the final step of the pipeline and shows how to re-assemble the local patches to a global reference system given the coarse 3D mesh. 

\subsection{Surface Patch Alignment}
\label{ssec:pimps:patch alignment}

A global and dense surface can be obtained only by registering each reconstructed surface patch to the corresponding position on the coarse 3D mesh $\mathcal{S}$.
Such mapping is possible through the inverse of transformations $\mat R_\mu$ in Eq. (\ref{eqn:flat individually}) to align all surface patches back to the 3D mesh.
In this way, we can map image pixels belonging to the registered triangle patch $\mat{\bar J}_{\mu}$ to the corresponding facet of the 3D mesh $\mathcal S$ as below:

\begin{equation}
\label{eqn:aligned image points to corresponding facet}
\mat{\tilde{P}}_{\mu} = \frac{1}{\zeta} \; \mat R_{\mu}^{-1} \; 
\begin{bmatrix}
		\vec x_{\mu}^\top \\
		\vec y_{\mu}^\top \\
		1
\end{bmatrix}
\end{equation}
where vectors $\vec x_{\mu}$ and $\vec y_{\mu}$ include $x$ and $y$ components of the pixels belonging to registered image patch $\mat{\bar J}_{\mu}$, and $\mat{\tilde{P}}_{\mu}$ is a matrix containing points of the corresponding facet of the 3D mesh.

Note that the transformed triangular plane is larger than the corresponding facet since we map the projection of enlarged triangles to the 3D mesh. These additional pixels help us to find the photo-geometric transformation since they are overlapping points along the borders of proximate triangular patches.
These overlapping points should be aligned with the corresponding ones on the neighbouring surface patches.
This constraint is used to obtain the final surface. 
We can represent the surface patch $\tilde{\mat S}_{\mu}$ on the 3D mesh as transformed image points from triangular image patch $\mat{\bar J}_{\mu}$ that are elevated along the facet normal $\vec N_f^{\mu}$ with respect to the recovered surface patch $\mat{\tilde{S}}_{\mu}$, where 

\begin{equation}
\label{eqn:surface patch}
	\mat{\tilde{S}}_{\mu} = 
	\begin{bmatrix}
		\vec{\tilde{s}}_{\mu 1} \; \ldots \; \vec{\tilde{s}}_{\mu \left|{\mathcal K_{\mu}}\right|}
	\end{bmatrix}_{3 \times \left|{\mathcal K_{\mu}}\right|},
\end{equation}
accumulates $\left|{\mathcal K_{\mu}}\right|$ 3D vectors representing points on the surface of triangular patch $\tilde{t}_{\mu}$. 

Therefore, the relation between the overlapping points for surface patches belonging to neighboring triangles $\tilde{t}_{\mu}$ and $\tilde{t}_{\mu'}$ can be described considering that the corresponding points from two surface patches should be identical on the global surface $\mat S$.
This relation is represented as:
\begin{equation}
\label{eqn:surface patch alignment}
\begin{array}{llr}
	\mat{\tilde{P}}_{\mu}^{(c)} + \left[ \vec 0_3 \; \vec 0_3 \; \vec N_f^{\mu} \right]_{3 \times 3} \; . \; \mat H_{\mu} \; . \; \mat{\tilde{S}}_{\mu}^{(c)}
	&=&\vspace{.2cm}\\
	\mat{\tilde{P}}_{\mu'}^{(c)} + \left[ \vec 0_3 \; \vec 0_3 \; \vec N_f^{\mu'} \right]_{3 \times 3} \; . \; \mat H_{\mu'} \; . \; \mat{\tilde{S}}_{\mu'}^{(c)},&&
\end{array}
\end{equation}
where $\mat{\tilde{P}}_{\mu}^{(c)}$ represents the common points between $\mat{\tilde{P}}_{\mu}$ and $\mat{\tilde{P}}_{\mu'}$, $\mat{\tilde{S}}_{\mu}^{(c)}$ give the corresponding points between $\mat{\tilde{S}}_{\mu}$ and $\mat{\tilde{S}}_{\mu'}$ and $\vec 0_3$ is a vector of $3$ zeros.
We can rewrite Eq. (\ref{eqn:surface patch alignment}) as: %(see appendix ??):
\begin{equation}
\label{eqn:surface patch alignment_kronecker}
\begin{array}{ll}
	vec \left( \mat{\tilde{P}}_{\mu}^{(c)} \right) + 
	\left( \left( \mat{\tilde{S}}_{\mu}^{(c)} \right)^\top \; \!\!\!\! \otimes \; \!\! \left[ \vec 0_3 \; \vec 0_3 \; \vec N_f^{\mu} \right] \right) 
	\; . \; vec \left(\mat H_{\mu} \right)
	=&\\ %\vspace{.2cm}\\
	vec \left( \mat{\tilde{P}}_{\mu'}^{(c)} \right) + 
	\left( \left( \mat{\tilde{S}}_{\mu'}^{(c)} \right)^\top \; \!\! \!\! \otimes \; \!\! \left[ \vec 0_3 \; \vec 0_3 \; \vec N_f^{\mu'} \right] \right) 
	\; . \; vec \left(\mat H_{\mu'} \right),&
\end{array}	
\end{equation}
where operator $\otimes$ is the \textit{Kronecker product}.

Writing Eq. (\ref{eqn:surface patch alignment_kronecker}) for all the points overlapping between every pair of triangles, the problem of estimating matrices $\mat H_{\mu}$ and $\mat H_{\mu'}$ can be formalized as an optimization problem:
\begin{equation}
\label{eqn:surface patch alignment_optimization problem}
\begin{array}{rl}
	&\min\limits_{\mat H_{\mu}, \mat H_{\mu'}} \left\| \!\!
	\begin{array}{c}
		\! \left( \left( \mat{\tilde{S}}_{\mu}^{(c)} \right)^\top \;\!\!\!\! \otimes \!\!\; \left[ \vec 0_3 \; \vec 0_3 \; \vec N_f^{\mu} \right] \right) \; . \; vec \left(\mat H_{\mu} \right) - \\
		\! \left( \left( \mat{\tilde{S}}_{\mu'}^{(c)} \right)^\top \;\!\!\!\! \otimes\!\! \; \left[ \vec 0_3 \; \vec 0_3 \; \vec N_f^{\mu'} \right] \right) \; . \; vec \left(\mat H_{\mu'} \right) + \\
		vec \left( \mat{\tilde{P}}_{\mu}^{(c)} \right) - vec \left( \mat{\tilde{P}}_{\mu'}^{(c)} \right) + k 
	\end{array} \!\!
	\right\|^2\\ \\
	&k = \frac{1}{\sum \left( C_{\mu} + C_{\mu'} \right)}
\end{array},
\end{equation}
where $C_{\mu}$ and $C_{\mu'}$ denote the surface curvatures for triangles $\tilde{t}_{\mu}$ and $\tilde{t}_{\mu'}$. 
The curvature term $k$ prevents the solutions for $\mat H_{\mu}$ and $\mat H_{\mu'}$ from being chosen in a way that flattens the surface patches $\mat{\tilde{S}}_{\mu}$.
Solving Eq. (\ref{eqn:surface patch alignment_optimization problem}) for the transformations $\mat H_{\mu}$ and $\mat H_{\mu'}$ with Least Squares aligns all the surface patches by solving their local bas-relief ambiguity up to an arbitrary reduced local ambiguity.

The global dense surface $\mathcal{\hat S}$ is obtained after variational refinement of the surface through minimizing the energy function: 

\begin{equation}
\label{eqn:variational refinement}
	E(\mathcal{\hat S}) = \int \lambda (1-G(\vec s)) \left\| \bigcup_{\mu} \mat S_{\mu} - \mathcal{\hat S} \right\|_2^2 + G(\vec s) \left\| \nabla \mathcal{\hat S} \right\|_2^2 \mathrm{d}\vec s,
\end{equation}
where $\vec s$ is a surface point, $\bigcup\mat S_{\mu}$ is the geometric superposition of surface patches, and $\lambda > 0$ is a parameter to control the smoothness.
$G(.)$ is the weighting function
\begin{equation}
\label{eqn:weighting function}
	G(\vec s) = \frac{1}{2}\sqrt{\left\| \alpha^{\vec s} - \beta^{\vec s} \right\|_2 + \left\| \alpha^{\vec s} - \gamma^{\vec s} \right\|_2 + \left\| \gamma^{\vec s} - \beta^{\vec s} \right\|_2},
\end{equation}
where $\alpha^{\vec s}$, $\beta^{\vec s}$ and $\gamma^{\vec s}$ are the barycentric coordinates of 3D point $\vec s$ with respect to the corresponding triangle.

\section{Experiments}
\label{sec:experiments}

The proposed 3D reconstruction pipeline was tested on several sequences from \textit{RobotDataset} \cite{aanaes:etal:2011}\footnote{\url{http://roboimagedata.imm.dtu.dk}}. 
The main feature of this dataset is the possibility to have a dense 3D ground truth obtained by structured light. 
This represents a step forward with respect to the previous evaluation protocols where experimental results were mostly assessed on a qualitative basis. 
In terms of PS reconstruction, the sequences offer several challenges: i) the setup uses LED lights for illumination and it is completely uncalibrated; ii) strong shadows affect the images; iii) the general object's surface properties depart from a Lambertian model, iv) the corresponding PS problem is also large scale in its nature. 
Each single image has a frame size of $1600 \times 1200$ pixels, therefore, each image sequence has a number of elements (pixels) in the order of $10^9$.

\begin{figure*}[!]
%	\captionsetup{aboveskip=5pt}
	\begin{center}
		\includegraphics[trim = 0mm 20mm 0mm 20mm, clip=true, width=\linewidth]{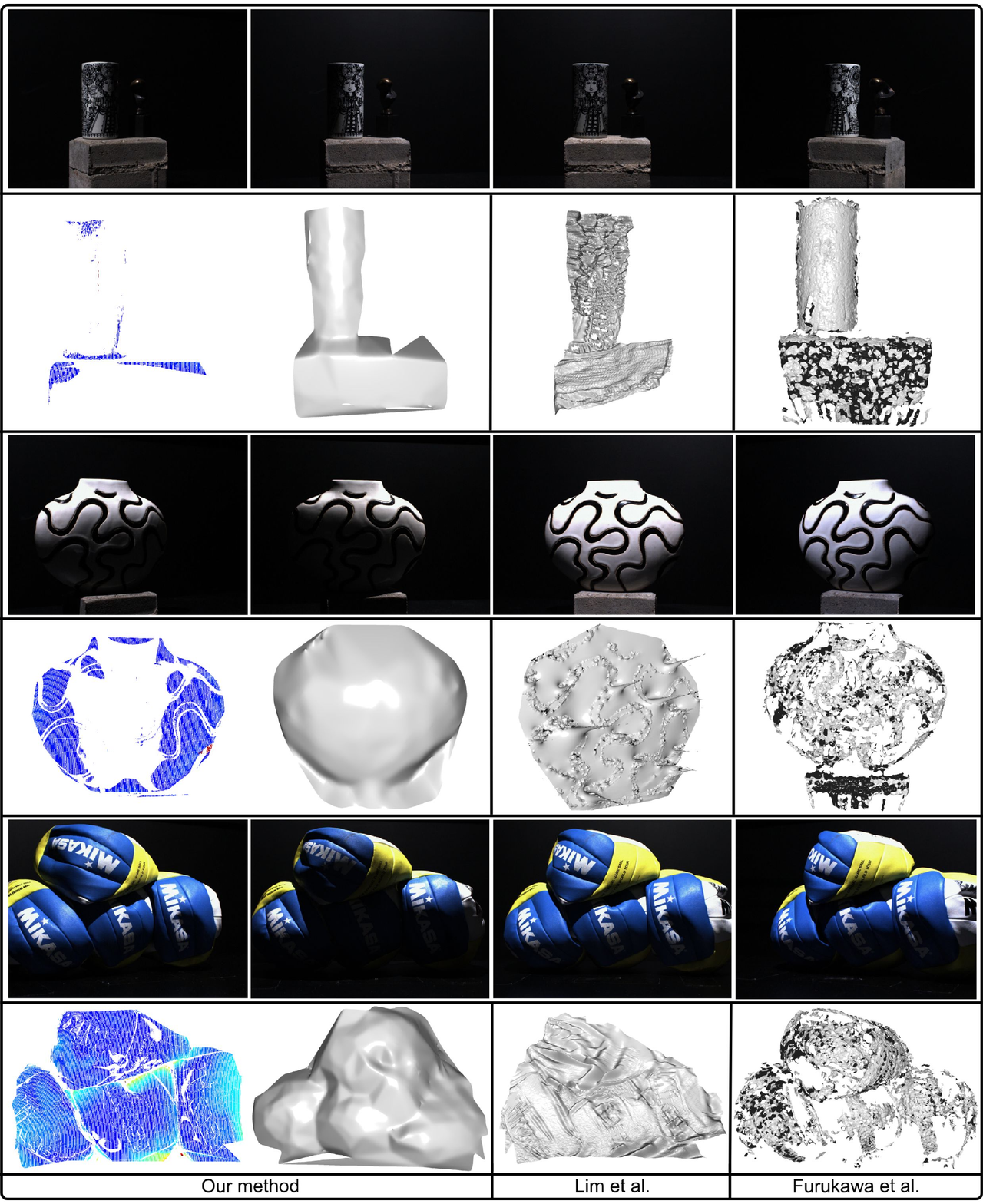}
	\end{center}
	\caption{Images from \textit{Mug}, \textit{Jar} and \textit{Ball} sequences together with our results (error heatmaps and 3D surface in the $1^{st}$ and $2^{nd}$ columns) and comparisons with a MVPS method \cite{lim2005passive} ($3^{rd}$ column) and a MVS method \cite{furukawa2010accurate}  ($4^{th}$ column).} 
	\label{fig:experiments}
\end{figure*}

\paragraph{Comparison with previous approaches} Given a test sequence in the dataset, we select $30$ frames with both moving camera and varying light sources. 
The camera motion on the arc gives a variation of $50$ degrees overall. 
In addition to the wide base-line, the images are high resolution and present strong shadows and reflection effects. 
Figure \ref{fig:experiments} shows the results for three sequences in the dataset (\emph{Mug}, \emph{Jar} and \emph{Ball}) compared with a classical MVS (i.e., Furukawa and Ponce \cite{furukawa2010accurate}), and the MVPS approach of Lim et al. \cite{lim2005passive}.  

The \emph{Mug} sequence has a glossy surface with a high reflectance but it is highly textured. 
Such properties make it challenging for standard PS but the texture is an advantage for MVS methods. 
In particular, this sequence shows that our approach can provide results comparable with MVS methods. 
The \emph{Ball} sequence has strong cast shadows and complex bending, which introduce interesting challenges for MVS methods. 
Although the surface is not reflective and can be considered as being almost Lambertian, the extreme lighting conditions give also a challenge to standard PS approaches. 
The \emph{Jar} sequence also has a glossy surface with  many saturated pixels which make the photometric reconstruction very challenging. 
Moreover, the smooth convexity and lack of sufficient texture introduce difficulties for MVS methods.

Multi-view photometric stereo approaches have the disadvantage of requiring short baseline images.
As a practical example, we have tested the method by Lim et al. \cite{lim2005passive}, which has similarities with our approach.
We have noticed that the main problem affecting this approach is the difficulty in finding correct pixel-to-pixel correspondences when the baseline between views increases. 
In most of the tested cases, the recovered shape by Lim et al \cite{lim2005passive} is almost flat with some bumps and few spikes (Figure~\ref{fig:experiments} - the 3rd column).

In contrast, multi-view stereo approaches can deal with wide baseline images but the reconstruction is not dense enough to include details of the surface.
The right column of Figure \ref{fig:experiments} (the 4th column) shows the reconstruction from Furukawa and Ponce \cite{furukawa2010accurate}.
Although the overall geometry of the recovered shape is acceptable, the surface can only be recovered for highly-textured parts of the shape, like in \emph{Mug} sequence.
The strong shadows in the \emph{Ball} sequence prevented this method from matching pixels on some parts of the surface, which is also the case in \emph{Jar} sequence because of poor texture.

\paragraph{Computational gain} Our piecewise formulation provides an efficient solution in terms of computational requests. 
This comes directly from considering several smaller PS problems instead of a single large-scale problem. 
In general, a piecewise formulation deals with multi-view patches of about $10^4$ pixels while a single PS example might amount to $10^8$ pixels approximately. 
To show such an effect, we compared the processing time and the performance for the \textit{potato} sequence example in \cite{del2012bilinear} using static view PS and an analogous sequence with our MVPS. 
The reported running time \cite{del2012bilinear} is almost $22$ hours and resulted in $17.34\%$ for mean 3D errors, instead our method achieved $11.15\%$ 3D error in $50.8$ minutes.

\vspace{.3cm}
{\setlength{\extrarowheight}{5pt}
\begin{table*}[bth]
	\caption{Quantitative data regarding our solution fo different sequences.}
	\label{tbl:results}
	\vspace{-.3cm}
	{\footnotesize
		\begin{center}
			\begin{tabular}
				{| >{\centering\arraybackslash}c | >{\centering\arraybackslash}c | >{\centering\arraybackslash}c | >{\centering\arraybackslash}c | >{\centering\arraybackslash}m{1.6cm} | >{\centering\arraybackslash}c | >{\centering\arraybackslash}c |}
				\hline

				Sequence 		& $\#$ Mesh Points 	& $\#$ Triangles & $\#$ Pixels 		& $\%$ Missing Pixels	& Time (Min.)	& 3D Error \\ 
				\hline
				Mug			& 115		& 198		& 6,689,940				& 24.7\%			& 24.38		& 7.1\%	\\	%& --\\
				\hline
				Jar			& 130		& 206		& 22,618,080			& 57.79\%			& 33.14 	& 9.34\% \\ %	& 9.56\% \\
				\hline
				Balls		& 250		& 362		& 42,374,490			& 69.48\%			& 62.63		& 9.69\% \\ %	& 9.72\% \\
				\hline
				Potatoes	& 150		& 276		& 36,672,990			& 62.72\%			& 50.8		& 11.15\% \\ %	& 11.18\%\\
				\hline
			\end{tabular}
		\end{center}
	}
\end{table*}}

Table \ref{tbl:results} provides some quantitative data regarding our solution. 
The rate of missing pixel values is noticeable and the obtained mean 3D error shows that the PPS could handle remarkable amounts of missing data. 
The timing presented in this table includes reconstruction of all the patches taking into account for the photo-geometric alignment.

\section{Conclusions}
\label{sec:conclusions}
We have presented a novel photo-geometric method for dense reconstruction from multi-view with arbitrary lighting condition.
The approach is able to cope with wide-baselines images from uncalibrated cameras and explicitly utilizes varying lighting directions in the image sequence.
This means that, only a few images taken by the end user are enough to be able to recover the dense 3D surfaces.
The piecewise approach is highly scalable since solving for image patches is computationally easier than considering whole images. 
This enables the pipeline to run on commodity PCs.
Future work will be dedicated to studying approaches that can partition the image into a mesh while taking into account the photometric and geometric properties of the shape. 
For instance, the method in \cite{Sunkavalli:etal:2010} could be used to partition the image in more consistent patches allocated to different subspaces. 
In addition, more complex photometric models could be used to extract more realistic photometric attributes for the surface.
%-------------------------------------------------------------------------
\section*{Acknowledgement}
This work is supported by the Doctoral fellowship awarded by the Italian Government and the Italian Institute of Technology.
The authors would like to thank Prof. Davide Scaramuzza for inspiring discussions and his valuable comments.

\bibliographystyle{ieeetr} 
\bibliography{pimps_ref}

\end{document}